\documentclass{article}

\usepackage{amsmath,amssymb,amsfonts}
\usepackage{algorithmic}
\usepackage{graphicx}
\usepackage{textcomp}
\usepackage{xcolor}
\usepackage{color}
\usepackage{todonotes}
\usepackage{footmisc}
\usepackage{array,etoolbox}
\usepackage{cleveref}

\newcounter{magicrownumbers}
\newcommand\rownumber{\stepcounter{magicrownumbers}\arabic{magicrownumbers}}
\preto\tabular{\setcounter{magicrownumbers}{0}}
\crefname{appsec}{Appendix}{Appendices}
\newcommand{\dataURL}{http://bit.ly/2DTkMh7}
\newcommand*{\fullref}[1]{\hyperref[{#1}]{\Cref*{#1}, \nameref*{#1}}}

\newcommand{\footremember}[2]{%
	\footnote{#2}
	\newcounter{#1}
	\setcounter{#1}{\value{footnote}}%
}
\newcommand{\footrecall}[1]{%
	\footnotemark[\value{#1}]%
} 

\title{Unsupervised Abbreviation Disambiguation \large \\ Contextual disambiguation using word embeddings}

\author{Manuel R. Ciosici \footremember{au}{Aarhus University, Denmark}\footremember{unsilo}{UNSILO A/S, Aarhus, Denmark}         \and
	Tobias Sommer\footrecall{au} \and
	Ira Assent\footrecall{au}}
\date{}
\usepackage[
style=numeric,
citestyle=numeric-comp,
natbib=true,
backend=biber,
]{biblatex}
\begin{document}

\maketitle

\begin{abstract}
Abbreviations often have several distinct meanings, often making their use in text ambiguous. Expanding them to their intended meaning in context is important for Machine Reading tasks such as document search, recommendation and question answering. Existing approaches mostly rely on manually labeled examples of abbreviations and their correct long-forms. Such data sets are costly to create and result in trained models with limited applicability and flexibility. Importantly, most current methods must be subjected to a full empirical evaluation in order to understand their limitations, which is cumbersome in practice.

In this paper, we present an entirely unsupervised abbreviation disambiguation method (called UAD) that picks up abbreviation definitions from unstructured text. Creating distinct tokens per meaning, we learn context representations as word vectors. We demonstrate how to further boost abbreviation disambiguation performance by obtaining better context representations using additional unstructured text. Our method is the first abbreviation disambiguation approach with a transparent model that allows performance analysis without requiring full-scale evaluation, making it highly relevant for real-world deployments. 

In our thorough empirical evaluation, UAD achieves high performance on large real-world data sets from different domains and outperforms both baseline and state-of-the-art methods. UAD scales well and supports thousands of abbreviations with multiple different meanings within a single model. 

In order to spur more research into abbreviation disambiguation, we publish a new data set, that we also use in our experiments.
\end{abstract}

\section{Introduction}
Abbreviations are shortened forms of phrases or single words, employed most often in written language. When the text denoting a concept is long and an author must refer to it multiple times, it is easier for the author to only use the abbreviation. Thus, abbreviations are many-to-one mappings from long-forms to short-forms, used when space or human limitations make the short-form more convenient. Due to the many-to-one mapping, abbreviation usage can cause problems for automated, computer-based, readers and represents a challenge for various natural language understanding tasks. For example, \emph{PCB} is an abbreviation for \emph{Polychlorinated biphenyl}, \emph{Printed Circuit Board}, \emph{Pakistan Cricket Board} and a few other concepts. Without domain knowledge, the meaning of \emph{PCB} in the sentence: \emph{The environmental measures taken include limiting the of substances such as cadmium, lead, PCB, and Azo pigments}, cannot be correctly inferred to be \emph{Polychlorinated biphenyl}. Abbreviations are particularly prevalent in biomedicine literature, often confusing readers \cite{Fred2003}. In MEDLINE abstracts, a high percentage of abbreviations can have multiple meanings ($64.6\%$) \cite{Liu2002a}. Structured knowledge bases such as the Unified Medical Language System (UMLS) also contain a considerable amount of ambiguous abbreviations (33.1\%) \cite{liu2001study}.

Resolving the intended meaning of an abbreviation in a given context is crucial for search engines, question answering systems, document recommendation systems, or text analytics. All systems addressing Natural Language Understanding tasks are expected to automatically know which concept a sentence refers to, even if only an abbreviation (i.e., short-form) is used to denote said concept. If a search engine is given a query like \emph{effects of Polychlorinated biphenyl on the environment}, one would expect the search engine to return documents referring to \emph{Polychlorinated biphenyl} either with its long-form or with its short-form. Similarly, recommender systems suggesting further literature on the topic of \emph{Printed Circuit Boards} should not suggest only documents where the authors employ exclusively the long-form, but also those where the short-form \emph{PCB} is used to denote the same meaning. Indeed, search engines like MEDLINE return different results when queried with abbreviations, corresponding ontology categories, or long-forms. This is the case even when in queries containing unambiguous abbreviations \cite{doi:10.1111/j.1553-2712.1999.tb00392.x}.

Previous abbreviation disambiguation methods have relied on corpora of manually annotated 
meanings and used hand-designed features \cite{moon2013word,moon2015challenges,wu2015clinical,wu2016long}. Such approaches have two major weaknesses. First, the cost of annotating abbreviation meanings is high as it involves human labor, often from domain experts. Secondly, models trained on such corpora can only disambiguate abbreviations that have been manually annotated. In order for such systems to support new abbreviations, more human work from domain experts is necessary to create more annotations, leading to higher costs. Similarly, if disambiguation is needed for a new domain, different domain experts are required to first manually annotate a corpus. Human annotation costs have prevented the scaling up of abbreviation disambiguation to large collections of abbreviations spanning multiple domains. Indeed, most data sets for abbreviation disambiguation are quite small \cite{li2015acronym,movshovitz2012alignment,bordes2011learning,wu2015clinical}.

Our method is completely unsupervised. It automatically extracts short-forms and their possible long-forms from large corpora of unstructured text. Thus it can learn what abbreviations exist and their possible meanings based on the provided unstructured corpus. We learn a different word vector for each long-form, and for all context words, in order to create semantic representations of the possible long-forms and the context in which they are used. Using this representation space, our method can easily identify the most likely intended long-form for an ambiguous abbreviation based on the context in which each abbreviation meaning is used. As the first method in this field, we provide an analysis of the expected disambiguation performance that does not require costly and cumbersome empirical evaluation studies on manually annotated test data. Additionally, we introduce a technique that can boost the quality of the context representations using external text corpora for learning. In summary, our method is completely unsupervised, generally applicable, flexible in terms of application domains and text content, as well as easy to review and update.

In this paper, we present the following contributions: (1) the Unsupervised Abbreviation Disambiguation (UAD), a disambiguation method that is unsupervised and does not employ hand-designed features, (2) a data set for evaluating performance on the abbreviation disambiguation task. For paper review purposes, the data set and evaluation scripts are available at \dataURL. They will be moved to the Harvard Open DataVerse before camera ready. 

\section{Related Work}
\label{sec:related_work}
Supervised approaches to abbreviation disambiguation require sense-annotated data sets, usually annotated manually, by domain experts. These methods often rely on hand-designed features such as co-occurrence (e.g., with other words in the corpus), number of uppercase letters etc \cite{moon2013word,moon2015challenges,wu2016long}. These features are usually designed using domain knowledge or are language-specific. Our method does not use hand-designed features. 

One approach to avoiding hand-designed features is to use word vectors to represent words and long-forms. A short paper by \citet{li2015acronym} briefly sketches this idea and presents two models built using word2vec vectors \cite{mikolov2013efficient}. Both models represent long-forms using a single vector calculated as a weighted sum of vectors of words observed in the context of each long-form, in all training examples. The two models differ in the underlying word2vec model and the weighing strategy. The first model, TBE, averages CBOW-derived word vectors of surrounding words using TF-IDF scores. It also filters the terms used in the average computation to keep only the top $n$ in the window surrounding a long-form usage. The other, SBE, represents context as the sum of vectors of contextual words observed in all occurrences in the training data, but uses Skip-gram-derived word vectors. Thus, for both models proposed by \citeauthor{li2015acronym}, the central idea is to create a fingerprint of the context in which each long-form is observed. Once fingerprint vectors have been computed to represent each long-form, abbreviation disambiguation reduces to picking that long-form candidate whose fingerprint vector has the smallest cosine distance to the average of vectors of words surrounding the abbreviation. \citeauthor{li2015acronym} show experimentally that the SBE model outperforms TBE. More recently, \citet{Charbonnier2018} present Distr. Sim., a variation of TBE that also uses context fingerprinting and  only differs from TBE in its choice of IDF as for term weighing instead of TF-IDF. Unlike these methods, in our model, vectors representing long-forms are not fingerprints of observed contexts, but trained in relation to their context, by a language model, at the same time as the rest of the vector space. Thus, their value is established by the language model underpinning word2vec, and values of long-form vectors influence those of contextual words through the language model used by word2vec. Furthermore, UAD does not employ context weighing strategies such as TF-IDF or IDF that require the extra step of computing corpus statistics.

One problem affecting unsupervised abbreviation disambiguation is that of identifying valid long-forms. Many unsupervised methods identify valid long-forms by first identifying definitions of abbreviations. Such methods must account for inconsistencies in writing long-form definitions. In the biomedical domain, \citet{okazaki2010building} found that many long-forms represent the same \emph{sense} even when lexically different, e.g., \emph{pathologic complete response} and \emph{pathologically complete responses}. They propose a normalization method using a supervised machine learning trained on a set of term variations manually annotated by domain experts. The classifier uses a number of hand-designed features like n-gram similarity measured using words and characters, number of shared letters, and a variation of TF-IDF. Disambiguation is based on hand-designed features built from unigrams and bigrams. Experiments show the value of normalizing long-forms and limiting lexical variations. In contrast, the normalization step used in our method is simpler and does not require any labeled data from domain experts.

There have been some attempts to formulate abbreviation disambiguation as a classification problem. Since the set of labels (i.e., long-forms) is specific to each short-form, \citet{wu2015clinical} express abbreviation disambiguation as a series of classification problems, one for each abbreviation (i.e., short-form). Their system combines word vectors with hand-designed features and uses a Support Vector Machine to train a classifier for each supported short-form. However, only distinct 75 abbreviations are included in the system. Such a method is impractical for industrial settings that require support for thousands of distinct abbreviations with, potentially, tens of thousands of long-forms. By contrast, our system builds a single model that allows disambiguation of all abbreviations and does not use hand-designed features, making it generally applicable, including for multiple languages.

Entity Linking (EL), often referred to as Record Linking, Entity Resolution or Entity Disambiguation, is a Natural Language Processing task that addresses ambiguities in text and bears some similarity to Abbreviation Disambiguation. The EL task consists of identifying entity mentions in text and providing links from the mentions to knowledge base entries. Entity Linking can target named entities (i.e., names that uniquely refer to entities of predefined classes such as person, location or organization), or common entities (also referred to as Wikification) where all noun phrases determining entities must be linked to concepts in a given knowledge base. The choice of entities to disambiguate depends on downstream tasks which can, for example, be to track named entities for document indexing, relation extraction, or question answering \cite{ling2015design,6823700}.

Both Abbreviation Disambiguation and Entity Linking establish connections between a lexical form and one of potentially many meanings. Despite this similarity, the two tasks differ in fundamental aspects. In Entity Linking ambiguity, at least in part, is due to partial forms which exist due to pre-established context (e.g., referring to \emph{George Washington} as \emph{George} or \emph{Washington}), metonymy (e.g., using the name of a country or its capital city as a substitute for its government), or entity overlap (e.g., mentioning a city together with its geographic location, such as \emph{Madison, Wisconsin}) \cite{TACL291,ling2015design,rao2013entity}. Often, in Entity Linking, the disambiguation task is simplified by the possibility of finding the correct linking via co-reference resolution.

Another core difference lies in the fact that Entity Linking is based on few distinct types (traditionally place, organization, person) that are insufficient to differentiate between different meanings of abbreviations \cite{rao2013entity}. More importantly, Entity Linking is a supervised task and requires ground truth, entity types for proper linking, as well as access to a knowledge base. In Abbreviation Disambiguation such types are not available and are not practical due to the large number of types and their granularity. Our method is entirely unsupervised, and does not require a knowledge base.

\section{Method}

In this section, we present Unsupervised Abbreviation Disambiguation (UAD), a method that requires only a corpus of unstructured text from which it extracts examples of abbreviation uses together with the intended meanings. Because of its simple input requirements, UAD can scale to large corpora and can easily be adapted to new domains by providing it with a corpus of unstructured text from the new domain. UAD exploits properties of \emph{word vectors} in order to represent the dependency between long-forms and the contexts in which they are used. UAD does not rely on hand-designed features.

Input to \emph{Abbreviation Disambiguation} can be formalized as s triple $(s_x, L_{s_x}, u)$, where $s_x$ represents a short-form, $L_{s_x}$ is the set of all known long-forms associated with short-form $s_x$, and $u$ is a sentence where the short-form $s_x$ is used without a definition. An abbreviation disambiguation system should output the long-form $l_y \in L_{s_x}$ that represents the meaning of $s_x$ intended by the author in the sentence $u$. In our experiments, we only focus on \emph{ambiguous abbreviations} (i.e., short-forms $s_x$ for which $|L_{s_x}| \ge 2$). We do not consider \emph{unambiguous abbreviations} (i.e., those $s_x$ for which $|L_{s_x}| = 1$. Unambiguous abbreviations can be expanded using a dictionary lookup as their expansion does not involve any form of disambiguation \cite{ciosici-assent-2018-abbreviation}. In the Unified Medical Language System, a popular ontology in the field of medicine, 76.9\% of abbreviations are unambiguous, and thus, require no disambiguation \cite{liu2001study}.

\subsection{Unsupervised Disambiguation}
\label{sec:method:disambiguation_method}
\begin{figure*}[t]
	\centering
	\includegraphics[width=1.0\textwidth]{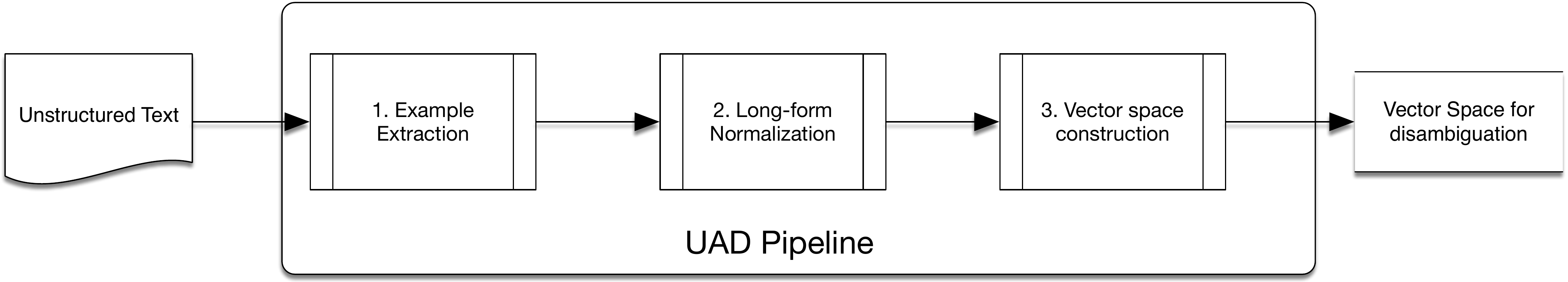}
	\caption{Overview of the UAD pipeline}
	\label{fig:uad_pipeline}
\end{figure*}

UAD is based on the idea of learning to represent words such that context can be used to predict the presence of often co-occurring words. We use this idea to express the disambiguation task as a prediction problem. An overview of the UAD pipeline is presented in \Cref{fig:uad_pipeline}. UAD's pipeline consumes unstructured text from which it extracts unambiguous examples of abbreviation uses together with the intended long-form. In the next step, lexical variations of long-forms are normalized, and the pipeline determines which abbreviations are ambiguous based on the count of normalized long-forms. Following, the set of examples containing ambiguous abbreviations with normalized long-forms is formulated as a text prediction problem and used for vector space construction. UAD's pipeline outputs a vector space containing vectors for both the various meanings of each recognized abbreviation and for all vocabulary terms observed in the corpus. We will now describe each state in detail.

\subsubsection{Example extraction}
\label{sec:example_extraction}

We start by splitting the large corpus of unstructured text into sentences and tokens using Stanford CoreNLP \footnote{\url{https://stanfordnlp.github.io/CoreNLP/}}, a common Natural Language Processing framework. We then extract sentences containing short-forms whose intended meaning is known (stage 1 in \Cref{fig:uad_pipeline}). We call these sentences \emph{unambiguous uses of abbreviations} since the intended expansion is defined locally in the text. Unlike previous work, we do not rely on human-annotated data sets to learn mappings. Instead, we extract them from text using an extension of the patterns defined by \citet{Schwartz2003}. The core idea expressed by the patterns is that abbreviations are often defined using the pattern \emph{long-form (short-form)} or \emph{short-form (long-form)}. For example, \emph{A printed circuit board (PCB) is a support for electronic components that also provides physical conductive connections} is an unambiguous use of the abbreviation \emph{PCB} meaning \emph{Printed Circuit Board}. From this sentence, we extract the tuple \emph{(\textless sentence\textgreater, PCB, Printed Circuit Board)}. The patterns also define character matching rules between the short-form and long-form that filter out most occurrences of the pattern that are not definitions. For example from the sentence ``Memory cards (SD) are used for data storage", is ignored as it contains no abbreviation definition. Our implementation of the long-form mapping implementation has an accuracy of $86.13\%$ on abbreviations in the Medstract data set \cite{Torii2007} and 62.48\% on those in Yeast \cite{Schwartz2003}. For details on the mapping algorithm and pseudocode, we refer the reader to the original paper \cite{Schwartz2003}. The performance difference between our implementation and the original is due to our stricter definition of abbreviations.

We identify abbreviations using regular expressions, and support single token abbreviations starting with a capital case letter and consisting of upper and lower case letters, digits, periods and dashes, but consisting of at least 60\% capital letters. Our implementation of the patterns proposed by \citeauthor{Schwartz2003} follows the description closely and thus, does not allow extra tokens inside parenthesis. As such, it cannot identify some abbreviation definitions such as ``Printed Circuit Boards (abbreviated PCB)" or ``Printed Circuit Boards (PCBs for short)". Missing such definitions is generally not a problem as abbreviations tend to be defined multiple times in large corpora. During this stage, we prioritize high extraction precision rather than recall. We do not focus on exhaustive extraction of all definitions since the meaning of abbreviations can be learned even if only a subset of each meaning's occurrences is extracted. Once an abbreviation definition is identified in a document, we assume that its meaning stays fixed in the document, and we also extract the subsequent uses in sentences as examples of usage for our model to learn from.

Thus, we apply a simple method to extract abbreviation usage examples from a large corpus. We do not aim for high extraction recall for definition occurrences, but rather on precision and try to eliminate wrong mapping extractions. Using a large corpus of text makes this strategy viable as abbreviations tend to be defined several times in a corpus, often with different styles. This stage of the UAD pipeline is designed as a module, so that it can be swapped with a better performing module when that is found.

\subsubsection{Long-form normalization}
\label{sec:normalization}

Long-form normalization (stage 2 in \Cref{fig:uad_pipeline}) follows example extraction and aims to filter noise and establish ambiguous abbreviations (i.e., short-forms that map into more than one long-form). We achieve both of these goals by normalizing long-forms that appear to denote the same meaning.

Long-forms that denote the same meaning can sometimes appear distinct due to inconsistent spelling (British vs. American English), different prepositions in long-forms, inconsistent use of hyphens, spaces, or plurals. For example, in the Wikipedia data, we found every single possible hyphenation of the long-form \emph{White Anglo-Saxon Protestants (WASP)}. To such inconsistencies, we normalize long-forms based on lexical similarity. We first strip long-forms of characters that cause lexical diversity without adding information such as spaces, hyphens, and ending characters that denote pluralism. This removes superficial variation in long-forms (e.g., \emph{Amino Acid} and \emph{amino-acids} are normalized to \emph{aminoacid}).

For each short-form, all stripped long-forms are compared with each other using the Levenshtein string edit distance \cite{Levenshtein}. We normalize the less frequent long-form to the more popular one, if the ratio between the string edit distance and the length of the longer long-form is low. This is based on the assumption that the correct spellings of long-forms are more popular than incorrect variations. Practically, we express each normalization decision as a rewrite rule so that for each tuple $(l_i, l_j)$, the normalization function $n$ returns that parameter to which both long-forms should be normalized, or \emph{null} to denote that no rewrite is necessary (i.e., the two long-forms are too distinct). The rewrite rules allow us to chain long-form rewrites so that, for example, we could normalize a long-form $l_i$ to $l_j$, and then $l_j$ to some other $l_h$ belonging to the same short-form (i.e., $l_i, l_j, l_h \in L_{s_x}$ for some $s_x$). Formally:
\begin{multline}
\forall s_x \text{ and } \forall (l_i, l_j) \in \lbrace L{s_x} \times L{s_x} \rbrace: \\ 
n(l_i, l_j)= 
\begin{cases}
\underset{\lbrace l_i, l_j \rbrace}{\arg\ \max}\ freq(l),& \text{if } \frac{lev(l_i, 
	l_j)}{\max(|l_i|,|l_j|)} \leq t\\
null,              & \text{otherwise}
\end{cases}
\end{multline}
where $|l_x|$ is the length of $l_x$. Threshold $t$ was set to 0.2 in order to 
keep rewrites conservative. Results $null$ are ignored.

After all rewrites are generated, we follow rewrite chains until we establish canonical long-forms. Normalization removes pseudo ambiguity in the data which allows us to (1) identify and ignore unambiguous abbreviations (i.e., those $s_x$ for which $|L_{s_x}| = 1$), (2) identify long-forms that denote distinct meanings, and (3) remove noise from the extraction stage as that can be easily identified as low-frequency mappings from short-forms to long-forms. UAD's normalization stage is simpler than the logistic classifier clustering used by \citet{okazaki2010building} which also requires manually labeled data from which to learn a similarity measure.

\subsubsection{Vector space construction}
\label{sec:vector_space_construction}

To disambiguate abbreviations, we formulate the task as a word prediction task where a model must predict the missing word in a sentence. Thus, in the vector space construction step (stage 3 in \Cref{fig:uad_pipeline}), we take the output from the previous step and reformulate each sentence. We replace all tokens constituting short-forms and their definitions by a single placeholder token representing the short-form and its intended, normalized, long-form. For example, the sentence mentioned above is rewritten to \emph{A \_\_ABB\_\_/PCB/printed\_circuit\_board is a support for electronic components that also provides physical conductive connections}. The special token \emph{\_\_ABB\_\_PCB/printed\_circuit\_board} allows us to identify the short-form that was used in the original sentence and its intended meaning. The \emph{\_\_ABB\_\_} prefix ensures that the placeholder token does not match any natural language token, and allows us to identify all short-forms and long-forms used in the data. In this example, the word prediction task becomes one of predicting that the special token \emph{\_\_ABB\_\_PCB/printed\_circuit\_board}, is the missing token in the sentence \emph{A \_\_\_\_ is a support for electronic components that also provides physical conductive connections}.

To train a model that can solve the prediction task, we turn to word vectors derived using language models, more specifically the method word2vec \cite{mikolov2013efficient} which derives word vectors that encode information from a language model using a shallow neural network. Word2vec can use two training strategies: \emph{Continuous Bag-Of-Words (CBOW)} and Skip-grams. When using CBOW, the probability of a word $w_i$ is estimated using information from a linear context constructed using a window of length $n$ on both sides of $w_i$ as:

$P(w_i | s ) = p(w_i | w_{i-1}, \dots, w_{i-n+1}, w_{i+1}, \dots, w_{i+n})
\label{eq:w2v_cbow_model}
$

When training using \emph{Skip-grams}, a word $w_i$ is used as input, and the model attempts to predict the surrounding words within a window of size $n$. In order to encode the language model, word2vec learns a representation of words in a multi-dimensional continuous space. Word representations can be used as inputs to the neural network in order to predict words according to the language model. Thus, most of the information required for language model prediction is encoded in the word vectors. Since the vectors are computed using a language model on a corpus of unlabeled text, word2vec is essentially unsupervised. However, word2vec requires many word repetitions in 
order to appropriately estimate a vector for each word \cite{mikolov2013efficient}.

We provide the sentences containing placeholder tokens to word2vec in order to derive word vectors that represent each token in the corpus (including the placeholders). We represent each long-form with a unique placeholder token so that, at training time, each long-form is allocated its own unique vector. The placeholder tokens are represented in relation to their contexts following word2vec's learning approach. Both placeholder tokens and words that often appear surrounding them, are represented in the vector space in such a way that their relationship is automatically encoded in the vectors using word2vec's underlying language model. In contrast, both SBE and Dist. Sim. rely on vector compositionality combined with weighing methods to construct representations for long-forms by creating fingerprints of the contexts observed in the data. They effectively ignore word2vec's underlying language model.

To disambiguate the use of a short-form $s_x$ in a sentence $u$, we first create a vector $c$ to represent the context around $s_x$ in $u$. The vector $c$ is the average of the vectors corresponding to words in the short-form's context. The disambiguation is the long-form candidate $l_i \in L_{s_x}$, which minimizes the cosine distance $m$ between the context vector $c$ and its own vector (i.e., the vector of the placeholder token): 
\begin{equation}
d(s_x, c)= \underset{l_i \in L_{s_x} 
}{\arg\ \min}\ m (c, l_i)
\end{equation}

Using this method, we can disambiguate any number of abbreviations without requiring data annotated by domain experts or hand-designed features. The only requirement is that each long-form is observed a sufficient number of times in the corpus before it can be reliably represented in the vector space. This requirement is not a limitation in practice as a larger unstructured corpus can always be found. The frequency requirement is also characteristic of all methods that rely on word vector spaces trained based on their usage. In our experiments we consider 50 observations to be sufficient for word2vec to construct stable and reliable vector representations. Since new short-forms and long-forms are introduced all the time, approaches relying on traditional, human-annotated data, require new manual annotations in order to support disambiguation of new abbreviations. UAD requires only a large corpus of unstructured text that contains uses with definitions of new abbreviations. Similarly, adaptation to a new domain only requires a large unstructured corpus of domain-specific text.

Global Vectors (GloVe) \cite{pennington2014glove} is a method for learning vector representations for words and is often considered an alternative to word2vec. It aims to combine the benefits of matrix factorization methods (i.e., the use of global corpus co-occurrence statistics, as opposed to only information that is available in the local window) with word2vec's skip-gram model in order to improve performance of resulting word vectors when used for word similarity, word analogy, or Named Entity Recognition. Both similarity and analogy tasks try to mimic the overall human perception of similarity between words or relations in pairs of words. In principle, GloVe vectors can be used as a drop-in replacement to word2vec in our disambiguation method. However, we expect that GloVe's prioritization of global co-occurrence to be detrimental to representations of long-forms as a function of context. The core idea in UAD is to allow a predictive language model to learn vector representations of both long-form meanings and words that appear in the context of abbreviations. For abbreviation disambiguation, we expect the words surrounding a short-form or long-form to be of much greater importance for correctly identifying the semantics than overall frequencies or co-occurrence as is the case for similarity and analogy tasks. In \Cref{sec:glove_vs_w2v}, we study experimentally GloVe as a drop-in replacement for word2vec as the word vector derivation method.

\subsection{Pre-evaluation analysis}
\label{sec:pre-evaluation_description}
In the previous section we presented UAD, an unsupervised method that can learn what abbreviations are present in a corpus, what meanings they have, and how to disambiguate them based on context. In this section, we present a property of UAD that addresses practical aspects of deploying abbreviation disambiguation in industrial settings.

Tailoring abbreviation disambiguation to a specific domain can improve performance by allowing the model to learn abbreviation meanings (i.e., long-forms) characteristic to the domain. Alternatively, one can avoid learning long-forms that are not used in the target domain, thus making disambiguation easier by reducing the number of long-form candidates for each short-form. For example, the \emph{Pakistan Cricket Board} long-form of the abbreviation \emph{PCB} used in examples earlier, is improbable to appear in chemical scientific literature. Automatically removing this long-form from the disambiguation model would eliminate the possibility of wrongly disambiguating to that long-form and, in some cases, it will reduce the number of possible long-forms for an abbreviation to 1, thus making the disambiguation process straight-forward (i.e., dictionary lookup). UAD is well suited for domain adaptation due to its unsupervised nature. 

Usually, after training a disambiguation model on a new corpus (or for a new domain), expensive, extensive evaluation is required to properly understand overall performance, and identify individual problematic cases, such as abbreviations whose long-forms are easily confused. Typically, this is done by repeated cycles of training, evaluation on test cases (for which costly manual labels are needed), followed by gathering, cleaning, and potentially labeling of new data. This iterative process can take considerable effort and time. Unfortunately, such tedious cycles are the norm for adapting existing methods to new domains \cite{moon2013word,wu2015clinical,wu2016long}. In industrial environments, where deployment speed of disambiguation models is crucial, costly, full-scale, evaluations on domain data may prove impractical. A method that does not require such expensive evaluation cycles can allow for faster and more flexible deployment in new domains, and for efficient model updates.

For UAD, we propose a novel, straightforward, rapid evaluation of the expected disambiguation quality, e.g., when updating to newly discovered abbreviations, or when tailoring the model to a new domain. Specifically, UAD can provide fully unsupervised insights into performance on target data, without requiring labeled data, or manual creation of test cases. Our pre-evaluation method is based on the observation that issues with abbreviation disambiguation performance translate directly into selecting incorrect long-forms for a given short-form and context. Consequently, we study which pairs of long-forms are difficult to disambiguate based on their context models. We show how this can be done using only the vector representations. If two long-forms belonging to the same short-form have similar representations in the vector space, they are clearly difficult to disambiguate. Let $l_i$ and $l_j$ be two long-forms belonging to the same short-form $s_x$, i.e., $l_i, l_j \in L_{s_x}$. If the vectors of $l_i$ and $l_j$ are similar under cosine distance, then any context vector $c$ that is close under the cosine distance to one of them, will also be close to the other. We only need to consider $l_i$ and $l_j$ which belong to the same short-form $s_x$ as UAD emits long-forms only from the set known to belong to the short-form that requires disambiguation. Thus, pairs of long-forms belonging to different short-forms do not cause problems for disambiguation. Formally, for the set $R_{s_x}$ containing all unordered pairs of long-forms that share the same short-form $s_x$: 
\begin{equation}
R_{s_x} = \lbrace \lbrace l_i, l_j \rbrace  \  | \ \exists \ s_x : l_i, l_j \in L_{s_x} \ 
\rbrace 
\end{equation}
we construct a ranking $P_{R_{s_x}}$ of the unordered pairs based on the cosine distance $m$ between the members of each pair. In other words, the unordered pair $\lbrace l_i, l_j \rbrace$ is more problematic than the unordered pair $\lbrace l_k, l_l \rbrace$ if and only if:
\begin{equation}
m(l_i, l_j) < m(l_k, l_l)
\label{eq:cosine_dist}
\end{equation}

In \Cref{eq:cosine_dist} we do not require that the two pairs of long-forms $(l_i, l_j)$ and $(l_k,l_l)$ belong to the same short-form $s_x$. The ranking $P_{R_{s_x}}$ can be easily reviewed for problematic long-form pairs without requiring costly evaluation on labeled test cases. The two unordered pairs of long-forms $\lbrace l_i, l_j \rbrace$ and $\lbrace l_k, l_l \rbrace$ need not belong to the same short-form as we are ordering pairs based on how difficult it is for UAD to distinguish between the elements of each pair, as opposed to between the pairs themselves. Our pre-evaluation based on the ranking $P_{R_{s_x}}$ only uses the trained disambiguation model and no labeled data. In \Cref{sec:experiments}, we measure the correlation between cosine distances of long-forms and disambiguation performance. Existing abbreviation disambiguation methods whose models are opaque to explanations \cite{wu2015clinical}, cannot provide such insights into their models, and thus must be subjected to a full-scale evaluation on ground truth, manually labeled data.

\subsection{Context Representations}
\label{sec:method:context_representations}
Proper context representation is critical to the performance of UAD. In the following subsections, we discuss how to create word vector spaces that effectively capture context.

\subsubsection{Augmentation with background text}
\label{sec:method:background_text}
As described previously, for derivation of vector spaces, UAD is provided with input consisting of only examples of unambiguous usage of ambiguous abbreviations that are automatically extracted from a large, unstructured, corpus. Small and targeted corpora may exhibit relatively high ratios of vocabulary to corpus length which means that words are observed relatively few times in the corpus. This can be problematic for unsupervised word representation methods such as word2vec. The small number of word uses may not provide enough information about the relation of words to their contexts, leading to non-representative vectors. In order to give word2vec a better chance to reach a stable vector space configuration, and to improve disambiguation performance, we augment the corpus used for vector space training. We supplement the set of abbreviation usage examples with an extra corpus of unstructured text. We aim for as high a vocabulary overlap with the abbreviation examples as possible in order to provide maximum word usage information. The addition of background text provides usage information for words that appear in both corpora. Since the location of long-form placeholders is influenced by that of all other words in the vocabulary through context-sharing, the locations of long-form placeholders will also be influenced, leading to overall improvements in abbreviation disambiguation. For maximal benefit, the augmentation text should share the same domain and writing style as the abbreviation examples. This can be achieved by extracting the background text from the same large, unstructured, corpus as the abbreviation examples. In experiments where we use background text, we denote such augmentations with \emph{TXT}.

\subsubsection{Stop-word removal}
\label{sec:method:stop_words}
Tokens that often occur in language use, such as prepositions, articles, punctuation etc. (often called stop-words) can reduce vector space quality. At training time, word2vec adjusts vectors of words surrounding stop-words in order to increase its performance in predicting stop-words. This behavior is undesired since it leads to lower quality vectors for tokens other than stop words. Moreover, our method (UAD) compounds this issue because at disambiguation time, it computes an average vector of all contextual tokens where content-bearing words have the same weight as stop-words.

In order to eliminate such undesired effects, we remove stop-words from the learning corpora. This also has the effect of effectively increasing the window size for word2vec as every stop-word removed allows another token to enter the window, if any is left in the sentence.

\section{Experiments}
\label{sec:experiments}
\subsection{Data Sets}
We used two data sets for all experiments: one based on the English Wikipedia, and one based on PubMed. All of English Wikipedia as of 1st of August 2017 was downloaded, and all text content was extracted using WikiExtractor \cite{wikiextractor}. For the PubMed data set, we downloaded the Commercial Use Collection of the Open Access Subset of PubMed \cite{pbmedopenaccess} and extracted only the text content from each article.

From each corpus, we selected only sentences with unambiguous abbreviation usage, and normalized long-forms using the methods described in \Cref{sec:example_extraction,sec:normalization}. We kept only those short-forms that had at least two long-forms (i.e., ambiguous abbreviations), each of which occurred in at least 50 examples. This minimum frequency threshold was chosen for two reasons. First, all three competitors considered in our experiments rely on vector space representations and are thus susceptible to unstable vectors for low-frequency terms. Second, the frequency threshold acts as a filter that reduces noise due to incorrect mappings during the first stage of our pipeline. All sentences containing uses of ambiguous abbreviations were distributed into 10 bins for the purpose of \emph{10-fold cross-validation}. Since we know the intended long-form for each of our extracted and normalized examples, we can perform \emph{10-fold cross-validation} without requiring manual labels. In order to provide more in-depth evaluation, we manually annotated a subset of the $10^{th}$ fold of our Wikipedia data set. This evaluation is discussed separately. The manually labeled data set is published together with the full Wikipedia data set and evaluation scripts.

For every data set, we also extracted a collection of sentences that is used as background corpus to derive higher quality word vectors (the background corpus is denoted TXT in all experiments).

A summary of the data sets is provided in \Cref{table:data_sets_overview}. We created the two data sets used in our experiments to address shortcomings in existing evaluations. Many data sets used in previous studies are too small to provide an accurate image of real-life disambiguation performance and some have artificial biases. For example, the MSH-WSD subset used by \citet{li2015acronym} has a low degree of ambiguity ($2.11$) and is heavily balanced, i.e., an almost equal number of usage examples are provided for each long-form. We do not consider such data sets sufficient for accurately assessing disambiguation performance. None the less, we have run experiments with the MSH-WSD subset from \citet{li2015acronym} for repeatability, which we include in \Cref{sec:competitor_comparisons} below. The data of \citet{Charbonnier2018} is not publicly available, but we have, of course, also evaluated their method, Distr. Sim. in our experimental study.

Even the smaller of our data sets, Wikipedia, is up to one order of magnitude larger than the ones used in previous literature \cite{li2015acronym,movshovitz2012alignment,bordes2011learning,wu2015clinical}. Our data sets help provide a better view of disambiguation performance as the data sets contain more varied test and learning examples. The ratio between the number of long-form examples in either of our data sets is not artificially balanced, i.e., they represent the relative popularity of the long-forms as observed in the corpora. The lack of artificial balancing allows for establishing a clear picture of the effectiveness of simple baselines, such as frequency-based disambiguation. We made the Wikipedia data set available online together with the evaluation script used in our experiments.

\begin{table}
	\centering
	\begin{tabular}{ r | r | r} 
		Dimension & Wikipedia & PubMed \\ [0.5ex] 
		\hline
		Examples (sentences) & $574\,837$ & $20\,812\,721$\\ 
		Short-forms & $567$ & $4\,139$ \\
		Long-forms & $1\,327$ & $17\,869$ \\
		Ambiguity & $2.34$ & $4.31$ \\
		Vocabulary size & $249\,293$ & $2\,836\,691$ \\
		Corpus length & $18\,238\,354$ & $756\,749\,229$ \\
		Vocabulary in TXT & $347\,658$ & $1\,472\,205$ \\
		Corpus length in TXT & $34\,326\,161$ & $256\,471\,477$ \\
		Voc. overlap with TXT & $54.10$\% & $34.47$\%\ \\
		Voc. overlap with GNEWS & $24.85$\% & $3.14$\%\ \\
		\hline
	\end{tabular}
	\caption{Overview of data sets. TXT refers to the background text created from each corpus. 
		GNEWS refers to the precomputed Google News vectors.}
	\label{table:data_sets_overview}
\end{table}

\subsection{Performance measures}
\label{sec:training_and_evaluation}

We measured performance as follows. For every long-form we calculated precision and recall as:
\begin{equation*}
\begin{aligned}[c]
precision = \frac{TP }{TP + FP}
\end{aligned}
\qquad\qquad
\begin{aligned}[c]
recall = \frac{TP }{TP + FN}
\end{aligned}
\end{equation*}

Where \emph{TP} stands for \emph{true positives}, \emph{FP} for \emph{false positives} and  \emph{FN} stands for \emph{false negatives}.

Performance measures are computed using micro-, macro-, and weighted versions of precision, recall and F1. For the purpose of evaluation, we treat the problem as multi-label classification. Thus, micro- versions of precision, recall, and F1 are equal to each other and take the same value as accuracy. Thus, we only report accuracy. Weights for the weighted measures are provided by the observed frequency of each long-form in the learning set. Overall scores are averages of each fold. We calculated both weighted and macro metrics in order to better understand the observed performance. Weighted precision, recall and F1 provide an estimation of real-life performance since abbreviations that occur more often have a higher weight. On the other hand, macro- measures describe how many abbreviations are properly disambiguated as there is no weighing between the different labels. For example, if an experiment leads to an increase in unweighted precision, but a drop in the weighted one, we can conclude that the new version handles less frequent abbreviations better, to the detriment of some frequent ones. The evaluation script that calculates all performance metrics is made available together with the Wikipedia data set.

\subsection{Results}
\label{sec:results}
\subsubsection{Setup}

UAD only requires those hyper-parameters specific to word2vec. More specifically, in all experiments we set \emph{w=10, i=10, size=300, neg=5} and the model to Skip-gram, unless stated differently.

\subsubsection{Comparisons with baseline and existing methods}
\label{sec:competitor_comparisons}

In the following experiments, we compare our method, UAD, with a simple, but efficient baseline, and two state-of-the-art methods: SBE \cite{li2015acronym} and Distr. Sim. \cite{Charbonnier2018}. The baseline (FREQUENCY) outputs the most frequent long-form for each short-form. Its statistics are calculated based on the frequencies observed in the 9 folds available for learning. Thus, the simple baseline FREQUENCY performs no context-based disambiguation and relies exclusively on corpus statistics. We implement SBE \cite{li2015acronym}, and its variation, Distr. Sim. \cite{Charbonnier2018}, and train both methods using the parameters given in the original papers. We adapt our pre-processing pipeline discussed \Cref{sec:normalization,sec:example_extraction} together with details in each of the competitors' papers in order to provide them with the same set of manually extracted examples as we use for our experiments. This evaluation is slightly biased against UAD as some competitors (SBE) are supervised methods requiring manually labeled data.

To evaluate SBE, \citet{li2015acronym} use a specifically constructed small subset of MSH WSD containing  abbreviations with little ambiguity (average ambiguity 2.11) and artificially balanced (the average ratio of long-form examples to total examples for each short-form is 1:2). On this data set, UAD outperforms both SBE \cite{li2015acronym} and Distr. Sim. \cite{Charbonnier2018}, see \Cref{table:competitor_comparison_MSH}. Due to the small size and artificial balancing of MSH WSD, we do not consider it to be representative for abbreviation disambiguation, and therefore proceed to evaluate on larger unbiased data.

\begin{table*}  
	\centering
	\begin{tabular}{ c | l | c | c | c | c | c | c | c  } 
		\multicolumn{3}{c|}{} & \multicolumn{3}{c|}{Weighted} & \multicolumn{3}{c}{Macro} \\[0.5ex]
		& Disambiguator & Acc. & Prec. & Rec. & F1 & Prec. & Rec. & F1 \\ [0.5ex] 
		\hline
		\rownumber & FREQUENCY & 54.14 & 30.04 & 54.14 & 38.46 & 25.55 & 46.34 & 32.79\\ 
		\rownumber & SBE \cite{li2015acronym} & 82.48 & 83.07 & 82.48 & 82.53 & 82.18 & 82.16 & 81.87 \\
		\rownumber & Distr. Sim. \cite{Charbonnier2018} & 80.19 & 80.87 & 80.19 & 80.25 & 79.90	& 80.12 & 79.71 \\
		\rownumber & UAD with TXT & \textbf{90.62} & \textbf{92.28} & \textbf{90.62} & \textbf{90.66} & \textbf{91.35} & \textbf{91.36} & \textbf{90.59}\\
		\hline
	\end{tabular}
	\caption{Competitor comparisons on the subset of MSH WSD used by \citet{li2015acronym}}
	\label{table:competitor_comparison_MSH}
	
\end{table*}

In \Cref{table:competitor_comparison} we show results of experimental comparisons on the Wikipedia and PubMed data sets. For both data sets, the FREQUENCY baseline manifests a discrepancy between the weighted and macro- measures. This is due to the disambiguation strategy which always disambiguates to the most popular long-form. Thus, it achieves perfect scores when disambiguating examples where the correct answer is the most popular long-form. Weighted performance is higher since less frequent long-forms matter less in that measure. SBE achieves performance results similar to what is reported in the original paper. Distr. Sim. is weaker than SBE in all measures. This is expected as Distr. Sim. is a variation of TBE, a weaker sibling of SBE, both proposed at the same time by \citet{li2015acronym}. Our model, UAD, outperforms the baseline and both competitors in all measures, on both data sets.

\subsubsection{Effect of background information}

As described earlier, we expected both data sets to exhibit relatively high ratios between vocabulary and corpus length. Indeed, in \Cref{table:data_sets_overview}, we can see that for the \emph{Wikipedia} data set, the ratio vocabulary size to corpus length is 1:73, while it is 1:266 for \emph{PubMed}. Given the relatively high ratio of vocabulary to corpus length, especially for the \emph{Wikipedia} data set, word2vec might not see each word in the vocabulary enough times in order to properly place each word in the vector space in relation to its context.

We evaluated the benefit of augmenting with background knowledge in two ways: directly and indirectly. The former consists of generating a background corpus of sentences extracted from the same data set as the examples of abbreviation use. Thus, it provides the same text style as the corpus of ambiguous abbreviation usage. The indirect method aims to study whether performance can be improved without increasing training time. For this, we used the GNEWS pre-computed vector space \cite{NIPS2013_5021}. The GNEWS space is trained on news items, so it does not share writing style with either of our corpora, but we consider it highly relevant since it was trained on a large corpus. In \Cref{table:data_sets_overview}, we show the vocabulary overlap of the learning text corpus (consisting of the examples extracted from \emph{Wikipedia} and \emph{PubMed}) with the background text (denoted TXT), and with GNEWS.

\begin{table*}  
	\centering
	\begin{tabular}{ c | l | c | c | c | c | c | c | c  } 
		\multicolumn{8}{c}{Wikipedia (no stop-words)}\\[0.5ex]
		\multicolumn{3}{c|}{} & \multicolumn{3}{c|}{Weighted} & \multicolumn{3}{c}{Macro} \\[0.5ex]
		& Disambiguator & Acc. & Prec. & Rec. & F1 & Prec. & Rec. & F1 \\ [0.5ex] 
		\hline
		
		\rownumber & FREQUENCY & 78.31 & 64.82 & 78.31 & 70.12 & 28.43 & 42.73 & 33.67\\ 
		\rownumber & SBE \cite{li2015acronym} & 89.54 & 93.40 & 89.54 & 90.79 & 84.92 & 88.43 & 85.41 \\
		\rownumber & Distr. Sim. \cite{Charbonnier2018} & 87.84 & 92.36 & 87.84 & 89.31 & 82.39 
		& 86.55 & 83.00 \\
		\rownumber & UAD with TXT & \textbf{94.28} & \textbf{96.17} & \textbf{94.28} & \textbf{94.76} & \textbf{90.84} & \textbf{93.29} & \textbf{90.98}\\
		\hline
	\end{tabular}
	
	\medskip
	\begin{tabular}{ c | l | c | c | c | c | c | c | c  } 
		\multicolumn{8}{c}{PubMed (no stop-words)}\\[0.5ex]
		\multicolumn{3}{c|}{} & \multicolumn{3}{c|}{Weighted} & \multicolumn{3}{c}{Macro} \\[0.5ex]
		& Disambiguator & Acc. & Prec. & Rec. & F1 & Prec. & Rec. & F1 \\ [0.5ex] 
		\hline
		
		\rownumber & FREQUENCY & 73.41 & 59.54 & 73.41 & 64.65 & 15.28 & 23.17 & 17.99\\ 
		\rownumber & SBE \cite{li2015acronym} & 69.33 & 89.08 & 69.33 & 74.96 & 67.84 & 78.88 & 69.27\\
		\rownumber & Distr. Sim. \cite{Charbonnier2018} & 62.94 & 87.21 & 62.94 & 69.39 & 61.19 & 73.85 & 62.32 \\
		\rownumber & UAD with TXT & \textbf{77.62} & \textbf{92.41} & \textbf{77.62} & \textbf{82.09} & \textbf{75.05} & \textbf{84.53} & \textbf{74.93}\\
		\hline
	\end{tabular}
	\caption{Baseline and competitor comparisons}
	\label{table:competitor_comparison}
	
	\bigskip 				\bigskip\bigskip
	\begin{tabular}{ c | l | c | c | c | c | c | c | c  } 
		\multicolumn{3}{c|}{} & \multicolumn{3}{c|}{Weighted} & \multicolumn{3}{c}{Macro} \\[0.5ex]
		& Augmented & Acc. & Prec. & Rec. & F1 & Prec. & Rec. & F1 \\ [0.5ex] 
		\hline
		\rownumber & No2 & 91.07 & \textbf{94.85} & 91.07 & \textbf{92.01} & 86.91 & \textbf{90.09} & \textbf{86.45}\\
		\rownumber & TXT2 & \textbf{91.10} & 94.78 & \textbf{91.10} & 91.91 & \textbf{87.61} & 89.56 & \textbf{86.45}\\
		\rownumber & GNEWS & 91.76 & 94.90 & 91.76 & 92.44 & 88.00 & 89.35 & 86.75\\
		\rownumber & TXT + GNEWS & 90.76 & 94.85 & 90.76 & 91.80 & 87.53 & 89.47 & 86.40\\
		\hline
	\end{tabular}
	\caption{UAD performance on the Wikipedia data set. w2v trained with \emph{model=Skip-gram, 
			w=10, i=10, 
			size=300, neg=5}}
	\label{table:wiki_background}
	\medskip
	\begin{tabular}{ c | l | c | c | c | c | c | c | c  } 
		\multicolumn{3}{c|}{} & \multicolumn{3}{c|}{Weighted} & \multicolumn{3}{c}{Macro} \\[0.5ex]
		& Augmented & Acc. & Prec. & Rec. & F1 & Prec. & Rec. & F1 \\ [0.5ex] 
		\hline
		\rownumber & No & 61.95 & \textbf{91.91} & 61.95 & 68.34 & 70.73 & \textbf{79.54} & 67.74\\
		\rownumber & TXT & 68.65 & 91.59 & 68.65 & 74.27 & \textbf{71.92} & 79.37 & \textbf{69.14}\\
		\rownumber & GNEWS & 62.50 & \textbf{91.91} & 62.50 & 68.86 & 70.80 & 79.53 & 67.78\\
		\rownumber & TXT + GNEWS & \textbf{69.11} & 91.49 & \textbf{69.11} & \textbf{74.51} & 71.90 & 78.48 & 68.43\\
		\hline
	\end{tabular}
	\caption{UAD performance on the PubMed data set. w2v trained with \emph{model=Skip-gram, 
			w=10, i=10, 
			size=300, neg=5}}
	\label{table:pubmed_background}
	\bigskip\bigskip\bigskip
	
\end{table*}
\begin{table*}  
	\centering
	\begin{tabular}{ c | l | c | c | c | c | c | c | c  } 
		\multicolumn{3}{c|}{} & \multicolumn{3}{c|}{Weighted} & \multicolumn{3}{c}{Macro} \\[0.5ex]
		& Augmented & Acc. & Prec. & Rec. & F1 & Prec. & Rec. & F1 \\ [0.5ex] 
		\hline
		\rownumber & No & 93.24 & 95.88 & 93.24 & 93.89 & 89.60 & 92.60 & 89.62\\
		\rownumber & TXT & 94.36 & 96.14 & 94.36 & 94.81 & \textbf{90.81} & 93.26 & \textbf{90.97}\\
		\rownumber & GNEWS & 93.26 & 95.86 & 93.26 & 93.89 & 89.57 & 92.61 & 89.61\\
		\rownumber & TXT + GNEWS & \textbf{94.37} & \textbf{96.15} & \textbf{94.37} & \textbf{94.82} & \textbf{90.81} & \textbf{93.27} & \textbf{90.97}\\
		\hline
	\end{tabular}
	\caption{UAD performance on Wikipedia without stop-words. w2v 
		trained with \emph{model=Skip-gram, w=10, i=10, 
			size=300, neg=5}}
	\label{table:wiki_background_no_stopwords}
	
	\medskip
	\begin{tabular}{ c | l | c | c | c | c | c | c | c  } 
		\multicolumn{3}{c|}{} & \multicolumn{3}{c|}{Weighted} & \multicolumn{3}{c}{Macro} \\[0.5ex]
		& Augmented & Acc. & Prec. & Rec. & F1 & Prec. & Rec. & F1 \\ [0.5ex] 
		\hline
		\rownumber & No & \textbf{78.03} & 92.41 & \textbf{78.03} & \textbf{82.28} & \textbf{75.29} & 84.56 & \textbf{75.17}\\
		\rownumber & TXT & 77.62 & 92.41 & 77.62 & 82.09 & 75.05 & 84.53 & 74.93\\
		\rownumber & GNEWS & 71.23 & \textbf{92.53} & 71.23 & 76.84 & 73.85 & 84.44 & 73.40\\
		\rownumber & TXT + GNEWS & 77.48 & 92.39 & 77.48 & 81.94 & 74.96 & \textbf{84.64} & 74.92\\
		\hline
	\end{tabular}
	\caption{UAD performance on PubMed without stop-words. w2v 
		trained with \emph{model=Skip-gram, w=10, i=10, size=300, neg=5}}
	\label{table:pubmed_background_no_stopwords}
	
	\bigskip
	
	\bigskip\bigskip	\bigskip\bigskip
	
	\begin{tabular}{c | c | c } 
		Data Set & Pearson Misclassification Rate & Spearman Misclassification Rate
		\\ [0.5ex] 
		\hline
		Wikipedia & $-0.73$ & $-0.43$\\
		PubMed & $-0.57$ & $-0.37$ \\
		\hline
	\end{tabular}
	\caption{Correlation between distance to closest other long-form and misclassification rate.}
	\label{table:correlation}
\end{table*}

Rows 2 of \Cref{table:wiki_background,table:pubmed_background} show that, indeed, addition of background text can improve performance. For Wikipedia a small increase can be observed in macro-recall indicating that background text improves performance on low-frequency abbreviations. For the other measures, we observe a small decrease in performance on the Wikipedia data set. In the case of the PubMed data set, accuracy, weighted recall, and macro-precision improve. Accuracy and weighted recall exhibit an improvement of 6.7 points.

Following, we initialized the vector space to the values in the GNEWS pre-computed vector space, and trained with our examples. Since GNEWS space is based on a large collection of news items, we expected initialization with this vector space to bring our own word vectors closer to the convergence point. Results for this experiment are presented in row 3 of \Cref{table:wiki_background,table:pubmed_background} (compare row 1 vs. 3 in both tables). For Wikipedia, just as before, macro-recall is improved, suggesting improved performance on low-frequency abbreviations. For PubMed, accuracy, weighted recall, and macro precision improve, while weighted precision and macro-recall are stable, or show little change. We believe the smaller improvements (and the drops in performance) are partly due to the lower vocabulary overlap with the GNEWS vectors (24.85\% and 3.4\%) compared to the background text (54.10\% and 34.47\%), see \Cref{table:data_sets_overview}.

Finally, rows 4 of \Cref{table:wiki_background,table:pubmed_background} contain results obtained when utilizing both kinds of augmentation. For both data sets, this leads to mixed results with the disambiguator benefiting more on the PubMed data set. In fact, comparing rows 2, 3, and 4 for PubMed, we can conclude that augmentation with background text is more effective than augmentation with GNEWS. The reverse is true for Wikipedia where seeding the vector space with GNEWS outperforms augmentation using background text. This might be due to the higher similarity of newswire text style to encyclopedic text than to scientific text.

The results of these experiments reveal an important aspect of our proposed unsupervised method: disambiguation performance can be improved by adding word usage information. Both augmentations are easy to employ as more vector spaces pre-computed over large corpora are becoming available.

\subsubsection{Effect of stop-word removal}

To evaluate the impact of stop-words, we re-processed our data sets and removed stop-words. A list of the eliminated tokens is available together with the Wikipedia data set and the evaluation script at \dataURL.

In \Cref{table:wiki_background_no_stopwords,table:pubmed_background_no_stopwords} we show the results of disambiguation on the two data sets after stop-word removal. For both data sets, the non-augmented vector spaces lead to disambiguation accuracy, precision, and recall that are higher than the augmented ones for spaces that contain stop-words. This is expected, given the way word2vec's underlying language model works, since the removal of stop-words eliminates noise from the window around each long-form. Results also show that it is highly important that the vector for each long-form is constructed only out of those words in the context that are semantically relevant. Furthermore, augmentation with background text or pre-trained vectors can still be applied and continue

\subsubsection{Comparison with Continuous Bag of Words}

word2vec supports two models for learning of word vectors: the Continuous Bag of Words Model (CBOW) and Skip-gram. In \Cref{table:cbow_wiki_background,table:cbow_pubmed_background,table:cbow_wiki_background_no_stopwords,table:cbow_pubmed_background_no_stopwords} we provide experiments using the CBOW model instead of Skip-gram. We use the same setup as our Skip-gram experiments in \Cref{table:wiki_background,table:pubmed_background,table:wiki_background_no_stopwords,table:pubmed_background_no_stopwords}.

UAD performs better when using the Skip-gram model compared to CBOW on both of our data sets. We believe this is due to the learning strategy in CBOW where a context vector is constructed and, from the context vector, a prediction is made. During training, the neural network assigns the prediction error to the context vector. Since it is not possible to determine which member of the context window is responsible for the error, the same correction is applied to the vectors of all words in the window. The Skip-gram model uses pairs of words and thus, vector corrections are applied proportionately to prediction error of each word's vector. Our experimental observation is in line with the original work of \citet{mikolov2013efficient}, who also conclude that word2vec with the Skip-gram model leads to more qualitative vector spaces.

However, some tendencies can be observed between our CBOW and Skip-gram experiments. First, addition of background data in the form of text or initialization with pre-trained vectors tends to improve disambiguation quality showing the importance of using more text to derive word vectors (e.g., row 2 in \Cref{table:cbow_wiki_background}). However, augmentation text from the same corpus generally leads to the highest disambiguation performance, probably due to matching text styles (e.g., in \Cref{table:cbow_wiki_background_no_stopwords} row 2 has a higher performance increase than row 3). Secondly, we observe that just as with Skip-gram, UAD based on CBOW performs worse on the PubMed data set, probably due to the more complex language (e.g., row 2 in \Cref{table:cbow_pubmed_background_no_stopwords} shows weaker disambiguation performance than row 2 in \Cref{table:cbow_wiki_background_no_stopwords}).

\begin{table*}
	\centering
	\begin{tabular}{ c | l | c | c | c | c | c | c | c  } 
		\multicolumn{3}{c|}{} & \multicolumn{3}{c|}{Weighted} & \multicolumn{3}{c}{Macro} \\[0.5ex]
		& Augmented & Acc. & Prec. & Rec. & F1 & Prec. & Rec. & F1 \\ [0.5ex] 
		\hline
		\rownumber & No & 77.91 & 92.30 & 77.91 & 80.42 & 81.23 & 85.24 & 80.24\\
		\rownumber & No2 & 76.72 & 92.13 & 76.72 & 79.36 & 80.69 & 84.89 & 79.53\\
		\rownumber & TXT & 79.31 & 92.47 & 79.31 & 82.25 & 82.00  & 85.49 & 80.94\\
		\rownumber & GNEWS & 78.05 & 92.33 & 78.05 & 80.70 & 81.31 & 85.29 & 78.05\\
		\rownumber & TXT + GNEWS & \textbf{79.46} & \textbf{92.51} & \textbf{79.46} & \textbf{82.37} & \textbf{81.99} & \textbf{85.56} & \textbf{80.99}\\
		\hline
	\end{tabular}
	\caption{UAD performance on the Wikipedia data set. w2v trained with 
		\emph{model=CBOW, 
			w=10, i=10, 
			size=300, neg=5}}
	\label{table:cbow_wiki_background}
	\medskip
	\begin{tabular}{ c | l | c | c | c | c | c | c | c  } 
		\multicolumn{3}{c|}{} & \multicolumn{3}{c|}{Weighted} & \multicolumn{3}{c}{Macro} \\[0.5ex]
		& Augmented & Acc. & Prec. & Rec. & F1 & Prec. & Rec. & F1 \\ [0.5ex] 
		\hline
		\rownumber & No & 69.31 & 86.82 & 69.31 & 74.18 & 61.69 & 69.31 & 60.24\\
		\rownumber & TXT & 69.78 & \textbf{86.89} & 69.78 & 74.60 & 61.77 & 69.24 & 60.30\\
		\rownumber & GNEWS & 68.99 & 86.85 & 68.99 & 73.94 & 61.62 & 69.32 & 60.15\\
		\rownumber & TXT + GNEWS & \textbf{69.94} & \textbf{86.89} & \textbf{69.94} & \textbf{74.75} & \textbf{61.78} & \textbf{69.40} & \textbf{60.43}\\
		\hline
	\end{tabular}
	\caption{UAD performance on the PubMed data set. w2v trained with 
		\emph{model=CBOW, 
			w=10, i=10, size=300, neg=5}}
	\label{table:cbow_pubmed_background}
	\bigskip\bigskip 		
	\begin{tabular}{ c | l | c | c | c | c | c | c | c  } 
		\multicolumn{3}{c|}{} & \multicolumn{3}{c|}{Weighted} & \multicolumn{3}{c}{Macro} \\[0.5ex]
		& Augmented & Acc. & Prec. & Rec. & F1 & Prec. & Rec. & F1 \\ [0.5ex] 
		\hline
		\rownumber & No & 86.30 & 94.12 & 86.30 & 88.64 & 86.07 & 90.06 & 86.30\\
		\rownumber & TXT & 87.80 & \textbf{94.43} & 87.80 & 89.88 & 87.34 & \textbf{90.70} & 87.54\\
		\rownumber & GNEWS & 86.30 & 94.15 & 86.30 & 88.64 & 86.13 & 90.08 & 86.20\\
		\rownumber & TXT + GNEWS & \textbf{87.93} & \textbf{94.43} & \textbf{87.93} & \textbf{89.98} & \textbf{87.36} & 90.68 & \textbf{87.55}\\
		\hline
	\end{tabular}
	\caption{UAD performance on the Wikipedia data set without stop-words. w2v 
		trained with \emph{model=CBOW, w=10, i=10, size=300, neg=5}}
	\label{table:cbow_wiki_background_no_stopwords}
	
	\medskip
	\begin{tabular}{ c | l | c | c | c | c | c | c | c  } 
		\multicolumn{3}{c|}{} & \multicolumn{3}{c|}{Weighted} & \multicolumn{3}{c}{Macro} \\[0.5ex]
		& Augmented & Acc. & Prec. & Rec. & F1 & Prec. & Rec. & F1 \\ [0.5ex] 
		\hline
		\rownumber & No & 74.69 & \textbf{88.94} & 74.69 & 78.94 & 67.31 & \textbf{77.37} & 68.06\\
		\rownumber & TXT & \textbf{75.74} & 88.89 & \textbf{75.74} & \textbf{79.66} & \textbf{67.78} & 77.14 & \textbf{68.51}\\
		\rownumber & GNEWS & 73.33 & 88.92 & 73.33 & 77.87 & 66.99 & 77.34 & 67.69\\
		\rownumber & TXT + GNEWS & 75.58 & 88.91 & 75.58 & 79.58 & 67.74 & 77.19 & 68.48\\
		\hline
	\end{tabular}
	\caption{UAD performance on the PubMed data set without stop-words. w2v 
		trained with \emph{model=CBOW, w=10, i=10, size=300, neg=5}}
	\label{table:cbow_pubmed_background_no_stopwords}
\end{table*}

\subsubsection{Comparisons between word2vec and GloVe}
\label{sec:glove_vs_w2v}
As discussed in \Cref{sec:vector_space_construction}, GloVe \cite{pennington2014glove} is a method for unsupervised construction of word vectors from large text corpora which attempts to combine word2vec's Skip-gram analogy capabilities with global corpus co-occurrence information. It is often considered an alternative to word2vec, especially for Named Entity Recognition, or tasks that involve similarity. 

In \Cref{table:glove_comparison}, we show a version of \Cref{table:competitor_comparison} with results from disambiguation with UAD, but using GloVe as a drop-in replacement for word2vec for construction of word vectors (see line 4). As mentioned earlier, we expect GloVe vectors to be less suited for abbreviation disambiguation, as the local context is of decisive importance for the representation of long-forms in the vector space. 

As expected, the experiments show that UAD using GloVe performs significantly worse than UAD using word2vec vectors. This observation is also in line with \citet{Charbonnier2018}, who also tested GloVe as part of Distr. Sim. and noticed a performance decrease. In fact, UAD with GloVe only outperforms the baseline FREQUENCY disambiguator. The results suggest that even though word2vec and GloVe are drop-in replacements for one-another in tasks that rely heavily on word similarity, that is not the case for abbreviation disambiguation, where local context is of high relevance.

\begin{table*}  
	\centering
	\begin{tabular}{ c | l | c | c | c | c | c | c | c  } 
		\multicolumn{8}{c}{Wikipedia (no stop-words)}\\[0.5ex]
		\multicolumn{3}{c|}{} & \multicolumn{3}{c|}{Weighted} & \multicolumn{3}{c}{Macro} \\[0.5ex]
		& Disambiguator & Acc. & Prec. & Rec. & F1 & Prec. & Rec. & F1 \\ [0.5ex] 
		\hline
		
		\rownumber & FREQUENCY & 78.31 & 64.82 & 78.31 & 70.12 & 28.43 & 42.73 & 33.67\\
		\rownumber & SBE \cite{li2015acronym} & 89.54 & 93.40 & 89.54 & 90.79 & 84.92 & 88.43 & 85.41 \\
		\rownumber & Distr. Sim. \cite{Charbonnier2018} & 87.84 & 92.36 & 87.84 & 89.31 & 82.39 
		& 86.55 & 83.00 \\
		\rownumber & UAD using GloVe & 82.23 & 89.43 & 82.23 & 82.90 & 74.50 & 74.05 & 68.99\\
		\rownumber & UAD with TXT & \textbf{94.28} & \textbf{96.17} & \textbf{94.28} & \textbf{94.76} & \textbf{90.84} & \textbf{93.29} & \textbf{90.98}\\
		\hline
	\end{tabular}
	\caption{Comparisons using GloVe on the Wikipedia data set with no stop-words.}
	\label{table:glove_comparison}
	
\end{table*}

\subsubsection{Evaluation against human-labeled data}
Both our learning and testing data are automatically extracted using the pipeline described in \Cref{sec:example_extraction,sec:normalization}. To evaluate the reliability of results on these data sets, we manually labeled $7\,000$ examples from one of the 10 folds from our Wikipedia data set. We then trained UAD using only the other 9 folds under the Skip-gram model with background text augmentation.

On the manually labeled subset, UAD achieves a weighted precision and recall of 97.62 and 95.18, respectively; macro- precision and recall of 93.24 and 94.09. These results are close to the ones in \Cref{table:wiki_background_no_stopwords}, which indicates that the automatically extracted data set is of high quality and confirms the results we presented earlier.

Since the labeled data set is much smaller than the ones automatically created, it better lends itself to detailed error analysis. We used 5 categories to classify each disambiguation error. Of the 337 errors, 2.33\% are due to multi-level abbreviations (e.g., \emph{Communist\_Party\_of\_the\_United\_States} vs. \emph{Communist\_Party\_USA}). 9.67\% are due to inconsistencies in long-forms that our normalization step cannot handle (e.g., \emph{Average Annual Daily Traffic} vs. \emph{Annual Average Daily Traffic}). A portion of 5.33\% represents language mismatches in long-forms that mean the same thing (e.g., \emph{Federation of Association Football} vs. \emph{F\'ed\'eration Internationale de Football Association}). The second-largest source of error is due to incorrect long-form mappings in our pipeline (e.g., \emph{Advanced Placement Program} instead of \emph{Advanced Placement}). We believe most of these errors can be solved in the future through improved pre-processing.

Of the remaining mistakes, many are made for long-forms that appear in sufficiently different contexts while only a minority represent difficult cases that have similar contexts such as \emph{American Broadcasting Company} versus \emph{Australian Broadcasting Corporation}.

\subsubsection{Pre-evaluation analysis}
\label{exp:understanding_vector_space}
In order to demonstrate that UAD supports efficient and cost-effective pre-evaluation analysis, we investigate the correlation between cosine distance of long-form pairs and their misclassification rate. For each of the data sets, we selected the models corresponding to row 2 in \Cref{table:wiki_background_no_stopwords,table:pubmed_background_no_stopwords}. For each pair of long-forms, we calculated the cosine distance and the misclassification rate between the two long-forms, and then both their \emph{Pearson} and \emph{Spearman} correlation coefficients. The \emph{Pearson Correlation Coefficient} $\rho_P$ is defined as:
\begin{equation}
\rho_P = \frac{\text{cov}(X,Y)}{\sigma_x \sigma_y}
\end{equation}
While the \emph{Spearman Correlation Coefficient} $\rho_S$ is defined as:
\begin{equation}
\rho_S = 1- {\frac {6 \sum d_i^2}{n(n^2 - 1)}}
\end{equation}
While Pearson evaluates the linear relationship between two variables and assumes normally distributed data, Spearman evaluates the monotonic relationship. For example, if an increase in one variable leads to an increase in the other, but the relationship is not linear, the Pearson correlation coefficient will show a weaker correlation, while Spearman will not be affected by the lack of linearity. We, therefore, measure both correlation coefficients in order to obtain a clearer picture. Analysis results are shown in \Cref{table:correlation}.

Both \emph{Pearson} and \emph{Spearman} correlation coefficients show that, on both data sets, there is a strong negative correlation between cosine distance and misclassification rate. In other words, the closer two long-forms are to each other, the more likely it is that UAD will have difficulties selecting the correct disambiguation. This follows the hypothesis presented in \Cref{sec:pre-evaluation_description}: if two long-forms have similar representatives, then a context vector aligned with one of them, will also align with the other. 

Pairs of close long-forms provide information as to how disambiguation performance can be 
improved. More specifically cases where long-forms represent the same meaning, but are considered different due to: lack of support for multi-level abbreviations(e.g., \emph{United States Geological Survey} and \emph{U.S. Geological Survey) for \emph{USGS}}, language mismatches, difficult edge cases for long-form normalization (e.g., words swapped around like in \emph{House Committee on Un-American Activities} and \emph{House Un-American Activities Committee} for \emph{HUAC}). Finally, we also observe examples from cases where more data is required for proper disambiguation due to long-forms that are difficult to disambiguate due to them denoting concepts in the same domains (e.g., \emph{Metropolitan Railway} and \emph{Midland Railway} or \emph{American Broadcasting Company} and \emph{Australian Broadcasting Corporation}). 

The pre-evaluation analysis feature of UAD is extremely useful in practice as it allows the identification of abbreviations that are potentially difficult to disambiguate without requiring expensive evaluations, such as the \emph{10-fold cross-validation} we performed in this article. Abbreviations that are difficult to disambiguate can either be removed from the models, investigated for potential pre-processing errors or noise. They can be used to target corpus acquisition towards gathering more usage examples for the specific abbreviations, which can lead to better representations for the long-forms and, thus, higher disambiguation performance. 

Finally, for pairs of long-forms that denote the same meaning, the model can either be retrained after normalizing the long-forms to a single lexical representation, the model can be updated directly by replacing the two long-form vectors, or it can be updated by removing one of the two members of a pair of problematic long-forms. In \cite{ciosici-naacl2019-demo}, we present a system to identify problematic long-forms which supports various vector space adjustment strategies. We show that vector space adjustment through removal or averaging of vectors leads to UAD performance improvements equal to those obtained when retraining.

\section{Conclusion}
We presented Unsupervised Abbreviation Disambiguation (UAD), a fully unsupervised method for abbreviation disambiguation that does not require hand-designed features or any labeled data. UAD automatically identifies abbreviations used in a large corpus of unstructured text, determines the potential long-form meanings, and constructs a model that can disambiguate abbreviations based on context.

UAD creates word vector spaces for representation of long-forms relative to their context, by introducing distinct tokens for each long-form identified. The relation between long-form representations and the context surrounding ambiguous abbreviations thus captures the information required for successful disambiguation.

Through a thorough empirical evaluation, we demonstrate that UAD outperforms a realistic baseline and state-of-the-art methods. Our evaluation is based on two data sets that are at least one order of magnitude larger than previously used in literature, are more ambiguous, have not been artificially balanced, and are more representative of real-world performance.

We also presented methods to further improve UAD's performance through augmentation with easy to acquire, ubiquitous background knowledge in the form of unstructured text, or pre-computed vector spaces. 

This the first method that supports insights into disambiguation performance without requiring full-scale evaluation through pre-evaluation analysis which can help identify problematic abbreviations, help target corpus acquisition, and in some cases allow for model adjustments that do not require retraining.

Both augmentation and pre-evaluation analysis make UAD highly relevant for real-world use, especially in domains with thousands of ambiguous abbreviations, or those that lack large sets of manually annotated data. 

As our contribution to the research community, we publish a data set containing abbreviation annotations, which is at least an order of magnitude larger than current similar resources. We hope the data set will support repeatability and further research in abbreviation disambiguation.

\printbibliography
\end{document}